\documentclass[conference]{IEEEtran}
\usepackage{times}

\usepackage[numbers]{natbib}
\usepackage{multicol}
\usepackage[bookmarks=true]{hyperref}
\usepackage{graphicx}
\usepackage{caption}
\usepackage{lipsum}
\usepackage{amsmath}
\usepackage{xcolor}
\usepackage{amssymb}
\usepackage{booktabs}
\usepackage{colortbl}
\definecolor{myblue}{RGB}{60,134,209}

\definecolor{plotblue}{RGB}{60,134,250}
\definecolor{plotgreen}{RGB}{70,200,100}
\definecolor{plotpurple}{RGB}{80,80,180}

\hypersetup{
    colorlinks=true,
    linkcolor=red,    
    citecolor=myblue,    
    filecolor=black,    
    urlcolor=black,
}

\begin{document}

\title{Learning Humanoid Locomotion with \\World Model Reconstruction}

\author{Wandong Sun\textsuperscript{1}, Long Chen\textsuperscript{1}, Yongbo Su\textsuperscript{1,2}, Baoshi Cao\textsuperscript{1,†}, Yang Liu\textsuperscript{1,†}, Zongwu Xie\textsuperscript{1}\\[0.5ex]
\textsuperscript{1}Harbin Institute of Technology \quad \textsuperscript{2}Tongji University \quad \textsuperscript{†}Equal Advising}



%

\twocolumn[{
\renewcommand\twocolumn[1][]{#1}
\maketitle
\begin{center}
    \captionsetup{type=figure}
    \includegraphics[width=\textwidth]{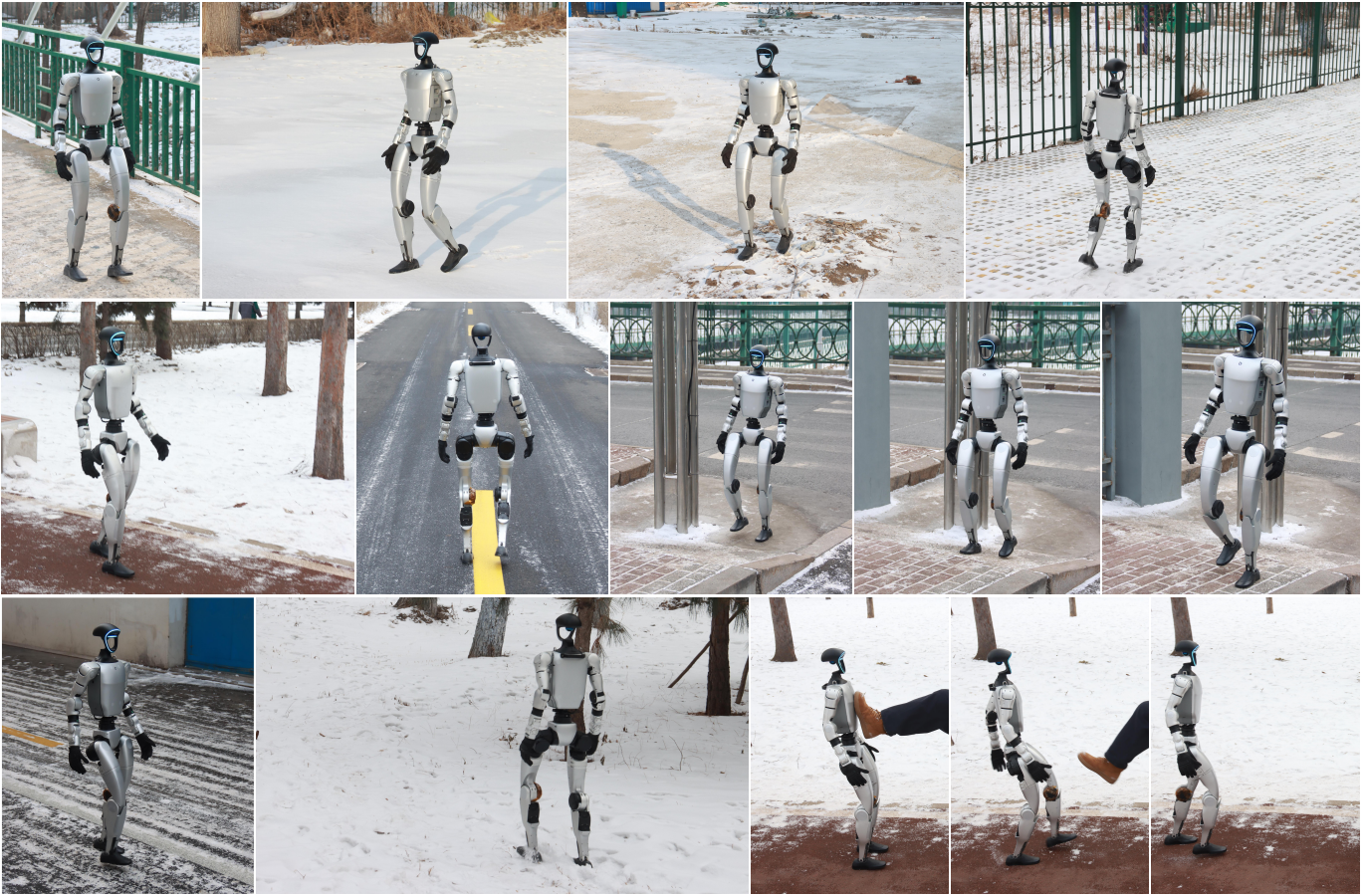}
    \captionof{figure}{\textbf{Deployment to outdoor environments.} We deployed the model in an outdoor environment covered in ice and snow. Our controller can successfully traverse a range of terrains, including rough, gravel, sloping, and deep snow terrains. It can also resist impact from humans and maintain stability on low-friction surfaces.}
    \label{fig:head}
\end{center}
}]

\begin{abstract}

Humanoid robots are designed to navigate environments accessible to humans using their legs. However, classical research has primarily focused on controlled laboratory settings, resulting in a gap in developing controllers for navigating complex real-world terrains. This challenge mainly arises from the limitations and noise in sensor data, which hinder the robot's understanding of itself and the environment. In this study, we introduce World Model Reconstruction (WMR), an end-to-end learning-based approach for blind humanoid locomotion across challenging terrains. We propose training an estimator to explicitly reconstruct the world state and utilize it to enhance the locomotion policy. The locomotion policy takes inputs entirely from the reconstructed information. The policy and the estimator are trained jointly; however, the gradient between them is intentionally cut off. This ensures that the estimator focuses solely on world reconstruction, independent of the locomotion policy's updates. We evaluated our model on rough, deformable, and slippery surfaces in real-world scenarios, demonstrating robust adaptability and resistance to interference. The robot successfully completed a 3.2 km hike without any human assistance, mastering terrains covered with ice and snow. We will release the code to completely reproduce the results presented in this paper.

\end{abstract}

\IEEEpeerreviewmaketitle

\section{Introduction}

\begin{table}[t]
    \centering
    \small
    \begin{tabular}{c c c c c}
        \toprule
        \textbf{Components} & \textbf{Dims} & \textbf{Proprio.} & \textbf{Recon.} & \textbf{Input} \\
        \midrule
        Angular Velocity & 3 & \checkmark & \checkmark &\checkmark \\
        Projected Gravity & 3 & \checkmark & \checkmark &\checkmark \\
        Command & 3 & \checkmark & \checkmark &\checkmark \\
        Joint Positions & 29 & \checkmark & \checkmark & \checkmark\\
        Joint Velocities & 29 & \checkmark & \checkmark & \checkmark\\
        Last Action & 29 & \checkmark & \checkmark & \checkmark\\
        \midrule
        Base velocity & 3 &   & \checkmark & \checkmark \\
        Contact mask & 2 &   & \checkmark & \checkmark \\
        Foot friction & 2 &   & \checkmark & \checkmark \\
        Payload & 1 &   & \checkmark & \checkmark \\
        Gravity & 1 &   & \checkmark & \checkmark \\
        Joint Stiffness & 29 &   & \checkmark & \checkmark \\
        Joint Damping & 29 &   & \checkmark & \checkmark \\
        Motor Offset & 29 &   & \checkmark & \checkmark \\
        \midrule
        Total & 192 & 96 & 192 & 192 \\
        \bottomrule
    \end{tabular}
    \caption{Summary of state space for sensing and reconstruction. The table categorizes the components into proprioceptive space, reconstruction space and policy input space.}
    \label{tab:observation}
\end{table}

Humanoid robots have the potential to perform tasks similar to humans by replicating human body structures and movements, based on the premise that these robots possess locomotion capabilities comparable to those of humans. This includes traversing natural environments such as dirt and gravel, as well as artificial settings like city parks, and handling slippery conditions involving ice and snow. However, developing robust controllers capable of managing a variety of challenging terrains remains a significant challenge.

Legged locomotion has been extensively studied, particularly in quadrupedal~\cite{lee2020learning,hoeller2024anymal,nahrendra2023dreamwaq,jenelten2024dtc,miki2022learning,kumar2021rma} and bipedal~\cite{kumar2022adapting,RealHumanoid2023,krishna2022linear,siekmann2021blind,li2024reinforcement} robots. Quadrupedal robots have made significant progress, demonstrating the ability to navigate a variety of challenging terrains with relative ease. In contrast, bipedal locomotion presents greater complexity due to inherent instability and dynamic balance requirements. Humanoid robots, which feature heavy legs and bulky upper bodies, face even more complex challenges in maintaining balance and achieving smooth locomotion.

Traditional humanoid locomotion control strategies, such as the zero-moment point~\cite{vukobratovic2004zero} principle and model predictive control~\cite{scianca2020mpc}, have enabled robots to perform reliable locomotion in structured environments~\cite{kuindersma2016optimization,sugihara2002real}. These approaches are typically extended to uneven terrains through complex step planning and sensory perceptual integration. Recently, reinforcement learning (RL) has emerged as a promising alternative to traditional control methods for legged robot locomotion~\cite{radosavovic2024humanoid,RealHumanoid2023,tang2024humanmimic}, leveraging the power of trial-and-error learning to develop adaptive and robust locomotion policies.

However, the design of both traditional and learning-based systems is significantly constrained by the available sensors, which ultimately impact the robot's locomotion capabilities. Crucially, the real world contains far richer information than what sensors can capture, and the captured information is often noisy. Can we effectively suppress noise and leverage seemingly unavailable information to enhance locomotion policies?

Here, we propose \textbf{World Model Reconstruction (WMR)}, a novel end-to-end reinforcement learning (RL) approach for mastering blind humanoid locomotion across challenging terrains. Our model comprises a world model estimator, a value network, and a locomotion policy. The estimator is trained to explicitly reconstruct the world state from sensor histories and supply it to the locomotion policy. The locomotion policy receives the estimated state and maps it to actions. The value network is trained to evaluate the policy. Our method employs a unified training procedure in which the estimator, value network, and locomotion policy are trained simultaneously. A critical aspect of our approach is the gradient cutoff between the estimator and the policy, ensuring that the estimator focuses exclusively on world reconstruction without being influenced by updates to the locomotion policy. Furthermore, we utilize a command space derived from motion capture dataset, fully exploiting the locomotion performance aligned with human distribution. We test our model in real-world scenarios. As shown in Fig.~\ref{fig:head}, our robot navigates stably through rough, deformable, sloping and slippery terrains. The contributions of our work are summarized as follows:

\begin{itemize}
    \item[1)] Present the first framework for augmenting humanoid locomotion with explicit world reconstruction, where the locomotion policy's inputs are entirely derived from explicit reconstruction.
    \item[2)] Propose a simple and effective gradient cut off mechanism, significantly enhancing the accuracy of the reconstruction.
    \item[3)] Utilize a command space derived from the motion capture dataset, combined with the WMR framework, to generate precise command tracking in line with human distribution.
\end{itemize}

\begin{figure*}[t]
    \centering
    \includegraphics[width=\textwidth]{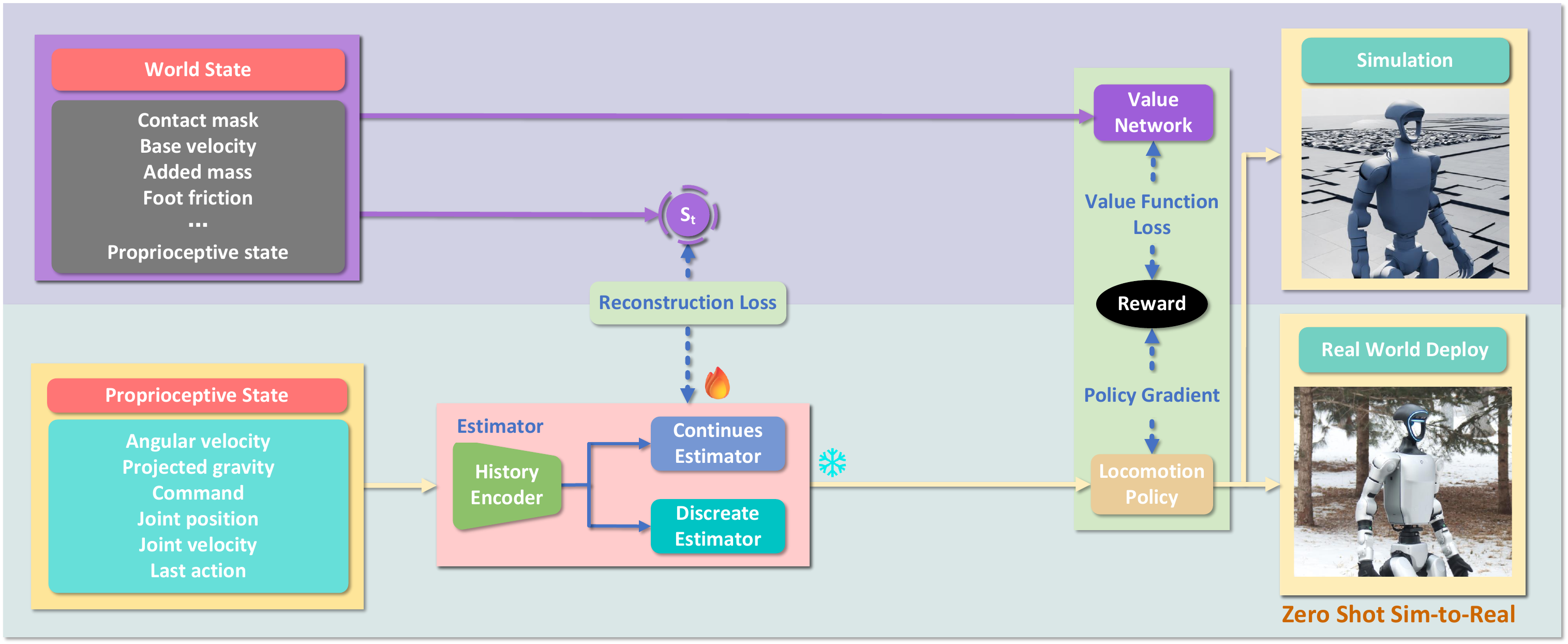}
    \caption{Illustration of the World Model Reconstruction framework. Our framework explicitly reconstructs world state from noisy sensor history and uses it as the sole input to the locomotion policy. The framework is driven by reconstruction loss, value function loss and policy gradient, innovatively applies gradient cutoff between the estimator and the locomotion policy to enhance the reconstruction accuracy. After one stage of training procedure, the robot can perform zero-shot sim-to-real transfer to challenging terrains.}
    \label{fig:method_overview}
\end{figure*}

\section{Related Works}

\subsection{Learning Humanoid Locomotion}
Reinforcement learning-based methods for legged robots have made great progress. Relying on a set of reward specification, legged robots can perform robust locomotion~\cite{jenelten2024dtc,lee2020learning,kumar2022adapting}, blindly~\cite{gu2024advancing,siekmann2021blind,nahrendra2023dreamwaq} and perceptively~\cite{long2024learning,chen2024identifying} traversing stairs and parkour~\cite{zhuang2024humanoid,hoeller2024anymal,zhuang2023robot}. These methods usually involve complex multi-stage training processes or behavior cloning, which may not work well in complex real-world situations and noisy sensor readings. Our framework is used for sensor denoising and world state estimation to minimize the sim-to-real gap.

\subsection{Learning Legged State Estimation}
Learning methods for state estimation of legged robots have been widely studied, with the main approach being to predict the privileged information from observation history. Previous work~\cite{kumar2021rma, kumar2022adapting} encoded the robot state and the environment state as latent variables and trained an adaptation module to imitate the encoded state. However, it requires multiple steps of training and the fully preservation of privileged information in the hidden state is not guaranteed. Previous work~\cite{ji2022concurrent} proposed a concurrent training method for the policy and the state estimator, which needs only one step of training with explicit estimation of the privileged information. Previous work~\cite{nahrendra2023dreamwaq} explicitly estimates the root speed and represents other environmental states as hidden states, training the decoder to reconstruct the privileged information. Previous work~\cite{gu2024advancing} went a step further and used the observation history to denoise the current observation and estimate the world state. The denoised state is stored in hidden variables.

However, these methods~\cite{nahrendra2023dreamwaq,gu2024advancing,ji2022concurrent} do not avoid passing the gradient of policy learning back to the estimation network, which will distract the goal of the estimation network and affect the estimation accuracy. In our work, we use sensor history to reconstruct the world state, which is used as the sole input of the policy. Due to the existence of gradient cutoff, there is a clear division of labor between modules, which greatly improves the denoising ability and estimation accuracy.

\section{Problem Formulation}

We formulate the problem of humanoid locomotion as goal-conditioned reinforcement learning within the context of a Markov Decision Process (MDP) to learn a policy $\pi: \Gamma \times \mathcal{O} \rightarrow \mathcal{A}$, where $\Gamma$ is the goal space to follow the commanded root velocity, $\mathcal{O}$ is the observation space and $\mathcal{A}$ is the action space, specifically the target joint positions to PD controller. We assume that both the observation and action spaces are defined according to the G1 humanoid robot design.

The locomotion goal is to track $\mathit{xy}$-plane linear velocity $\boldsymbol{v}_t$ and yaw rate $\Delta y_t$ from the pelvis coordinate system. The observation space is detailed in Table~\ref{tab:observation}, which contains the goal commands and information including joint encoder and IMU readings $\boldsymbol{o}_t = \left\{ \boldsymbol{\omega}_t, \boldsymbol{p}_t, \boldsymbol{v}_t, \Delta y_t, \boldsymbol{q}_t, \dot{\boldsymbol{q}}_t, \boldsymbol{a}_{t-1} \right\}$, where $\boldsymbol{\omega}_t$ is the angular velocity, $\boldsymbol{p}_t$ is the projected gravity vector, $\boldsymbol{q}_t$ is the joint positions, $\dot{\boldsymbol{q}}_t$ is the joint velocities, and $\boldsymbol{a}_{t-1}$ is the last action with default joint bias. The action output by the policy, $\boldsymbol{a}_t$, is then converted into motor torque through a PD controller.

\section{Methods}
We present \textbf{World Model Reconstruction (WMR)}, as shown in Fig.~\ref{fig:method_overview}, an end-to-end approach for privileged information reconstruction and sensor denoising, which is used to generate robust humanoid locomotion in real-world scenarios. In the following sections, we cover the key components of this approach.

\subsection{World Model Reconstruction}

\subsubsection{Context-Aided Estimator}

We propose training a state estimator to accurately recover sensor data and reconstruct privileged information from noisy sensor histories. Building upon the methodology presented by~\cite{nahrendra2023dreamwaq}, we employ domain randomization (DR) to simulate noisy sensor readings and utilize a context-aided estimator that effectively leverages historical observations to reconstruct the world state. Detailed domain randomization settings are provided in Appendix~\hyperref[subsec:append_dr]{B}. The estimator is optimized to minimize the reconstruction loss $\mathcal{L}_{\text{recon}}$, which quantifies the discrepancy between the predicted and actual world states $\boldsymbol{S}_t$. Architecturally, the estimator adopts an encoder-decoder framework, where a single-head encoder processes the perceptual history $\boldsymbol{O}_t^m$ to extract features, and a multi-head decoder reconstructs the comprehensive world model. This structure ensures that all decoders operate on a unified and consistent feature representation, facilitating better coordination among different reconstruction tasks~\cite{shazeer2019fast} and leading to more accurate estimations of the world state. All reconstruction information is shown in Table~\ref{tab:observation}.

The reconstructed information encompasses both denoised sensor data and explicitly estimated privileged information, providing the motion control strategy with direct, rich, and unadulterated inputs. Recognizing that the world state consists of both continuous and discrete components, we partition the decoder into two specialized branches: a continuous decoder and a discrete decoder. The continuous decoder is optimized using the Mean Squared Error (MSE) loss to reconstruct the continuous aspects of the world state:

\begin{equation}
\mathcal{L}_{\text{MSE}} = \frac{1}{N} \sum_{i=1}^{N} \| \mathbf{c}_i - \hat{\mathbf{c}}_i \|^2
\end{equation}

Conversely, the discrete decoder incorporates a sigmoid activation function in its final layer and is trained using the Binary Cross-Entropy (BCE) loss to enhance contact mask estimation:

\begin{equation}
\mathcal{L}_{\text{BCE}} = -\frac{1}{N} \sum_{i=1}^{N} \left[ y_i \log(\hat{y}_i) + (1 - y_i) \log(1 - \hat{y}_i) \right]
\end{equation}

To further improve the efficiency and robustness of state estimation, we introduce an L1 regularization term on the latent representation, leveraging its inherent sparsity~\cite{chen2022sparsity}:

\begin{equation}
\mathcal{L}_{\text{L1}} = \sum_{j=1}^{M} |z_j|
\end{equation}

The overall reconstruction loss for the estimator is thus a combination of these components:

\begin{equation}
\mathcal{L}_{\text{recon}} = \lambda_{\mathrm{cont}}\mathcal{L}_{\text{MSE}} + \lambda_{\mathrm{dis}}\mathcal{L}_{\text{BCE}} + \lambda_{\mathrm{reg}} \mathcal{L}_{\text{L1}}
\end{equation}

where $\lambda_{\mathrm{cont}}$, $\lambda_{\mathrm{dis}}$, and $\lambda_{\mathrm{reg}}$ are coefficients that control the weighting of each term. By estimating privileged information such as ground friction coefficients, base velocity, and contact masks, the robot can perform closed-loop command tracking and achieve robust adaptation to varying environmental conditions.

\subsubsection{Policy Learning}
We employ an asymmetric actor-critic architecture for policy learning, leveraging privileged information to enhance learning efficiency. The value network receives noise-free world information, while the locomotion policy takes its input from the world reconstruction provided by the context estimator. The locomotion policy is defined as $\pi(a_t|P_\text{estimator}(s_t|o_t^m))$, mapping reconstructed states to joint actions. The locomotion policy is updated using Proximal Policy Optimization (PPO)~\cite{schulman2017proximal}, which applies multiple steps of stochastic gradient descent (SGD) to maximize the expected return. The loss function is defined as:

\begin{equation}
    \begin{aligned}
        L_\pi = &\min \left(\frac{\pi_\theta(a|s)}{\pi_{\theta_k}(a|s)}A^{\pi_{\theta_k}}(s, a), \right. \\
        &\left.\text{clip}\left(\frac{\pi_\theta(a|s)}{\pi_{\theta_k}(a|s)}, 1-\epsilon, 1+\epsilon\right)A^{\pi_{\theta_k}}(s, a)\right)
    \end{aligned}
\end{equation}

where $A^{\pi_{\theta_k}}(s, a)$ is the advantage function, $\epsilon$ is the clipping parameter, and $\pi_{\theta_k}$ represents the policy from the previous iteration. The advantage function is computed using the value network $V_\phi(s)$, which is trained to minimize the error between the predicted and actual returns. The update rule for the value network is given by:

\begin{equation}
L_v = \left(V_\phi(s_t) - \hat{R}_t\right)^2
\end{equation}

where $\hat{R}_t$ is the actual return, and $V_\phi(s_t)$ is the predicted return from the value network.

\subsubsection{Loss Formulation}

Asymmetric policy learning integrates seamlessly with our world state reconstruction by employing a single optimizer to simultaneously train the estimator, value network, and locomotion policy, thereby ensuring synchronized updates and efficient training. The overall loss of the system is expressed as:

\begin{equation}
L = \mathcal{L}_{\text{recon}} + \lambda_v L_v + \lambda_\pi L_\pi
\end{equation}

where $\lambda_v$ and $\lambda_\pi$ are coefficients that modulate the relative contributions of the value network loss and policy loss. This formulation enables the motion strategy to dynamically adapt to the reconstructed state information from the estimator. As the motion strategy improves, it generates higher quality trajectories that provide more optimal input to the estimation network. This creates a positive feedback loop between the two components, leading to concurrent improvements throughout the training process and resulting in a more robust and efficient integrated system.

\begin{table*}[t]
    \centering
    \small
    \begin{tabular*}{\textwidth}{@{\extracolsep{\fill}} l ccccc}
        \toprule
        & \multicolumn{5}{c}{\textbf{Metrics}} \\
        \cmidrule{2-6}
        \textbf{Comparisons with Baselines} & $E_{\text{vel}} \downarrow$ & $E_{\text{ang}} \downarrow$ & $E_{\text{recon}} \downarrow$ & $M_{\text{terrain}} \downarrow$ & $M_{\text{reward}} \downarrow$ \\
        \midrule
        WMR without Gradient Cutoff & 0.693 & 0.421 & 0.459 & 0.569 & 5.123 \\
        WMR with Random Command Sample & 0.183 & 0.295 & 0.106 & 5.702 & 16.918 \\
        PPO & 0.318 & 0.364 & - & 4.695 & 13.892 \\
        Denoising World Model Learning & 0.250 & 0.310 & 0.154 & 5.596 & 15.156 \\
        \midrule
        World Model Reconstruction (Ours) & \textbf{0.156} & \textbf{0.252} & \textbf{0.098} & \textbf{5.827} & \textbf{18.695} \\
        \bottomrule
    \end{tabular*}
    \caption{Comparisons of metrics between baselines. We sample 1,000 trajectories with 4096 environments in simulation and report their mean episode metrics.}
    \label{tab:baselines}
\end{table*}

\begin{figure}[t]
    \centering
    \includegraphics[width=0.43\textwidth]{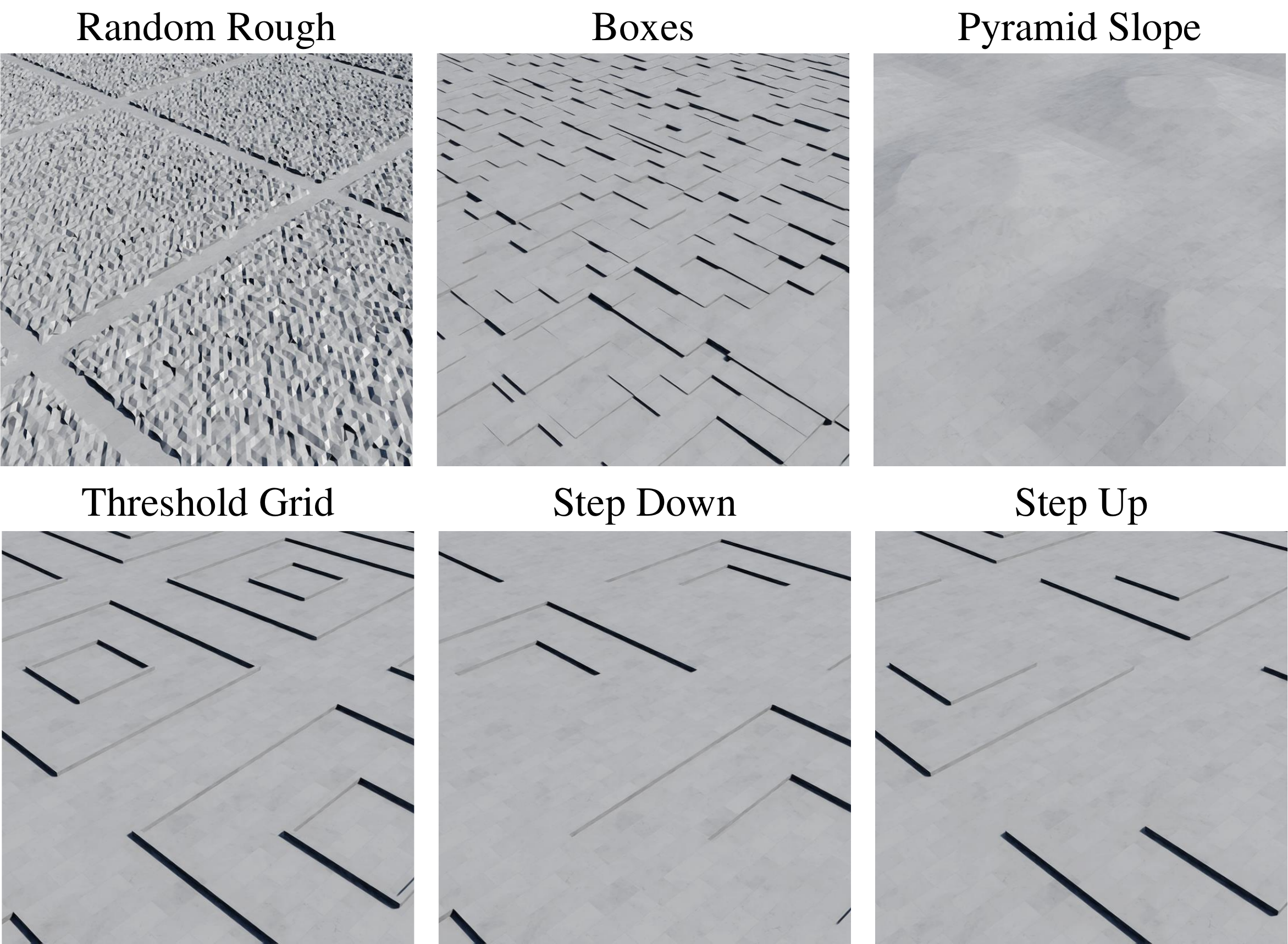}
    \caption{The robot is trained over a variety of terrains with random friction and restitution.}
    \label{fig:terrains}
\end{figure}

\subsection{Reward Formulation}

To fully unleash the robot's potential, we deliberately omit any predefined motion references in the reward functions. This includes, but is not limited to, periodic ground contact~\cite{gu2024humanoid,RealHumanoid2023,HumanoidTerrain2024}, foot lift height~\cite{cheng2024expressive,gu2024humanoid}, reference trajectory tracking~\cite{gu2024humanoid}, and stylized imitation~\cite{peng2021amp}. The reward function is composed of a task reward and a regularization reward~\cite{van2024revisiting}. The task reward is designed to encourage the robot to accurately track the commanded root velocity and yaw rate, while the regularization reward promotes the maintenance of a stable and human-like standing and walking posture. This results in a robust bipedal standing and walking controller that automatically switches cadence on command. Detailed specifications of the reward function can be found in Appendix~\hyperref[subsec:append_reward]{C}.

\begin{figure}[t]
    \centering
    \includegraphics[width=0.489\textwidth]{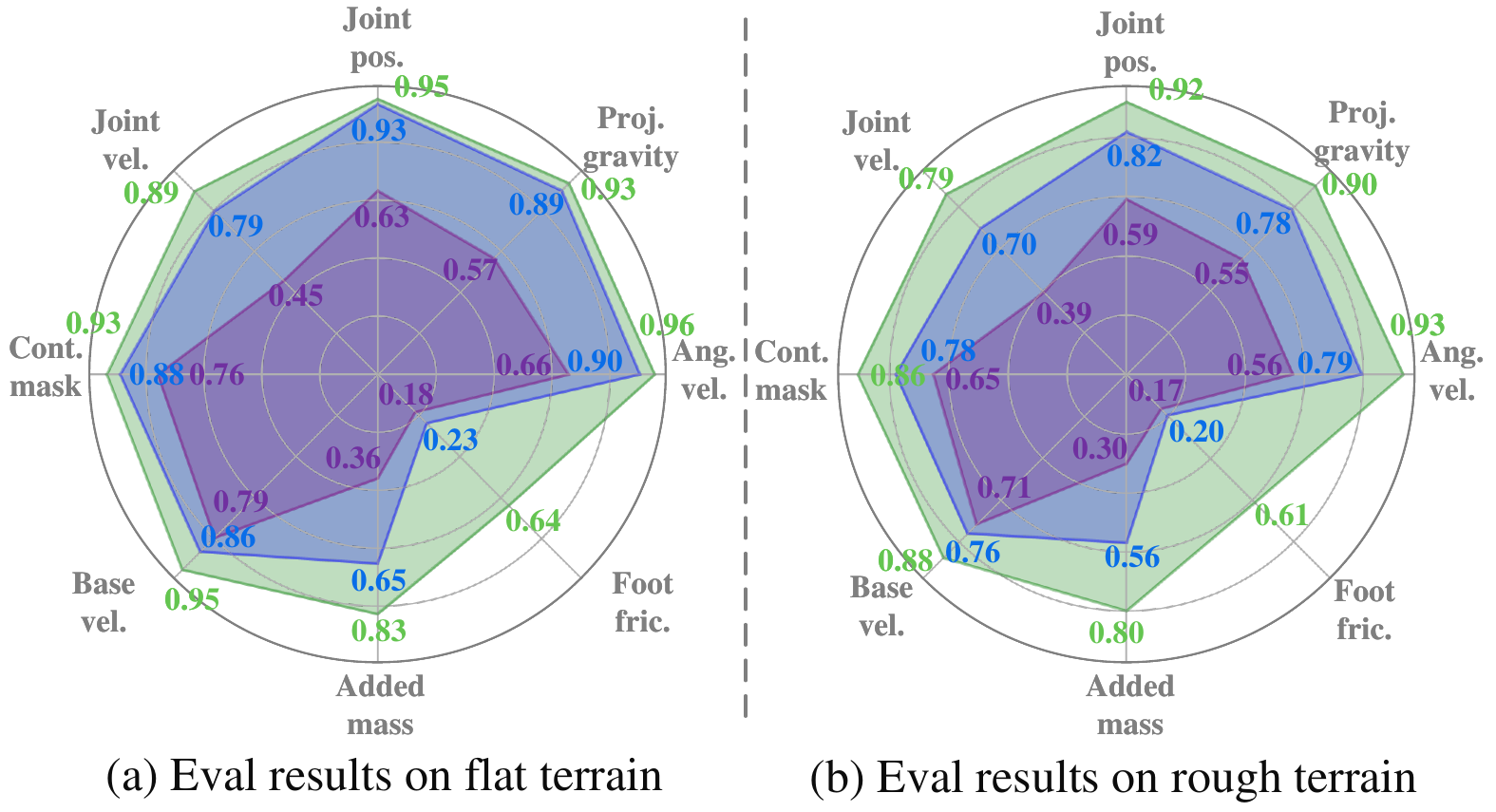}
    \caption{We eval the baselines on the reconstruction accuracy of the world state. The radar plot shows the mean reconstruction accuracy of each metric. The \textcolor{plotgreen}{green area} represents the WMR framework, the \textcolor{plotblue}{blue area} represents the Denoising World Model Learning framework, and the \textcolor{plotpurple}{purple area} represents the WMR framework without gradient cutoff.}
    \label{fig:radar_plot}
\end{figure}

\subsection{Terrain Curriculum}
The robot starts training on flat ground and adjusts the difficulty based on the length of the walk. The terrain, as shown in Fig.~\ref{fig:terrains}, includes a variety of terrains such as random rough, box obstacles, pyramid-shaped slopes, threshold barriers, and stairs.

\subsection{Command Configuration}
Our goal space, which encompasses the commanded linear and angular velocities, is sourced from the AMASS motion capture dataset~\cite{AMASS:2019}, as opposed to being obtained through random sampling in simulation. Specifically, we only utilize the root velocity extracted from the trajectories, eliminating the need for a retargeting process. A significant distinction between the motion capture dataset and random sampling lies in the distribution command space. The command space derived from the motion capture dataset more closely aligns with the motion distribution of humans. It features more natural speed transitions compared to random sampling, which is characterized by a fixed command. Moreover, allowing the robot to switch between different datasets facilitates its adaptation to the variations in the command space, thereby enhancing its versatility and robustness in responding to diverse motion requirements.

\subsection{Training Configuration}

\subsubsection{Model Architecture}
The estimator comprises an encoder and a multi-head decoder. The history encoder is implemented using an LSTM, which processes historical observations to generate a latent representation. The hidden and cell states of the LSTM propagate the robot's state across reasoning steps. The decoding module features multiple parallel MLP heads, each comprising two fully-connected layers with ELU activation

Additionally, both the locomotion policy and the value function are equipped with LSTM layers with the same hidden dimension as the encoder to effectively handle historical sequences of data. They are followed by MLP that connect to their respective output layers. The detailed network architecture is illustrated in Table~\ref{tab:network_arch}.

\begin{table*}[t]
    \centering
    \small
    \begin{tabular*}{\textwidth}{@{\extracolsep{\fill}} l ccccc}
        \toprule
        & \multicolumn{5}{c}{\textbf{Metrics}} \\
        \cmidrule{2-6}
        \textbf{Architecture Ablation} & $E_{\text{vel}} \downarrow$ & $E_{\text{ang}} \downarrow$ & $E_{\text{recon}} \downarrow$ & $M_{\text{terrain}} \downarrow$ & $M_{\text{reward}} \downarrow$ \\
        \midrule
        WMR-GRU-Memory & 0.162 & 0.259 & 0.156 & 5.764 & 18.231 \\
        WMR-MLP-History-25 & 0.198 & 0.295 & 0.226 & 5.326 & 16.893 \\
        WMR-Transformer-History-25 & \textbf{0.149} & 0.268 & 0.114 & 5.691 & 17.964 \\
        \midrule
        WMR-LSTM-Memory (Ours) & 0.156 & \textbf{0.252} & \textbf{0.098} & \textbf{5.827} & \textbf{18.695} \\

        \bottomrule
    \end{tabular*}
    \caption{Self Ablation Study: Evaluation of different history length and architecture. We sample 2,000 trajectories with 4096 environments in simulation and report their mean episode metrics.}
    \label{tab:Ablation}
\end{table*}

\subsubsection{Gradient Cut Off}
The gradient cutoff between the estimator and the locomotion policy is crucial to the success of our approach. This ensures that when computing \( L_\pi \), gradients are not back-propagated to the estimator, resulting in a clear division of responsibilities between the estimator and the locomotion policy. Without gradient cutoff, we observed a significant performance gap due to interference from competing gradients, which led to the estimator diverging from accurately reconstructing the desired world state. Consequently, the estimator's output would lose its interpretability and serve only as an intermediate product of the policy. Our evaluation demonstrates the necessity of the gradient cutoff within the WMR framework.

\begin{figure}[t]
    \centering
    \includegraphics[width=0.489\textwidth]{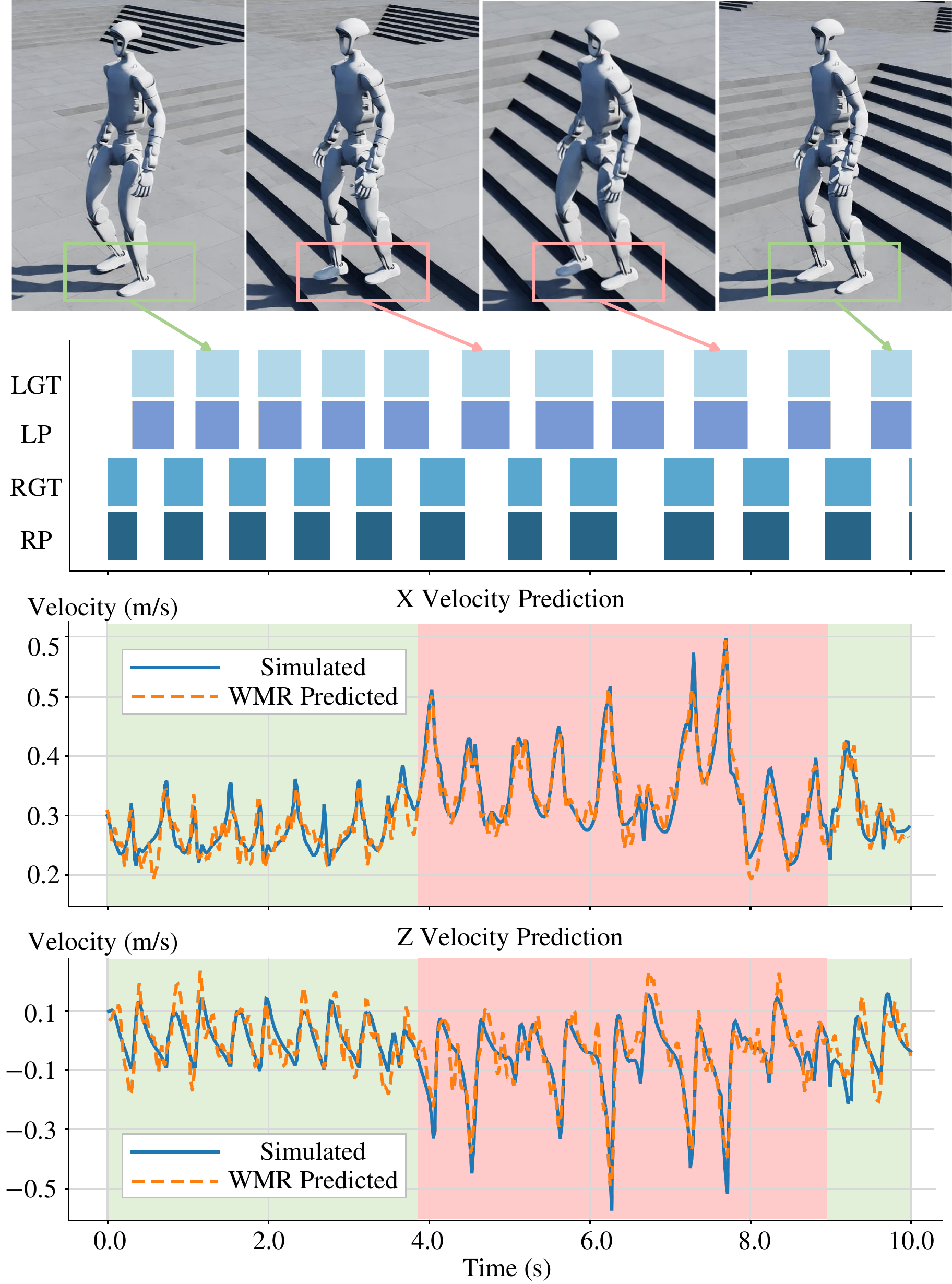}
    \caption{Comparison of ground truth and reconstruction. The ground truth data comes from the noise-free data in the simulator. The WMR framework receives the noisy data and predicts the world state.}
    \label{fig:est_sim}
\end{figure}

\section{Results}

In this section we aim to answer the following questions through extensive experiments both in sim and the real world:
\begin{itemize}
    \item How does the WMR framework perform on reconstruction accuracy?
    \item How much improvement in performance does the estimator provide?
    \item How does the gradient cutoff affect the performance?
    \item How does the architecture choice of the WMR framework affect the performance?
    \item How does the WMR framework perform on sim-to-real transfer on challenging terrains?
\end{itemize}

Our baselines are as follows:
\begin{itemize}
    \item \textbf{World Model Reconstruction (ours)}: This baseline is our proposed WMR framework with LSTM architecture and memory. The network architecture is shown in Table~\ref{tab:network_arch}.
    \item \textbf{WMR without Gradient Cutoff}: This baseline is our proposed WMR framework without gradient cutoff, which is used to evaluate the effectiveness of gradient cutoff. 
    \item \textbf{WMR with Random Command Sample}: This baseline is our proposed WMR framework with random command sample, which is used to evaluate the effectiveness of the motion capture command space.
    \item \textbf{PPO}: This baseline is the PPO framework with LSTM memory, which is used to evaluate the effectiveness of the world reconstruction.
    \item \textbf{Denoising World Model Learning}~\cite{gu2024humanoid}: This baseline is the denoising world model learning framework with encoder-decoder structure, which is seen as one of the state-of-the-art method in the field.
\end{itemize}

Our episode metrics are as follows:
\begin{itemize}
    \item \textbf{Mean Episode Linear Velocity Tracking Error} $E_{\text{vel}}$
    \item \textbf{Mean Episode Angular Velocity Tracking Error} $E_{\text{ang}}$
    \item \textbf{Mean Reconstruction Error} $E_{\text{recon}}$
    \item \textbf{Mean Terrain Levels} $M_{\text{terrain}}$
    \item \textbf{Mean Episode Reward} $M_{\text{reward}}$
\end{itemize}

\subsection{Comparisons between WMR and Baselines}
\noindent\textbf{Comparisons on Reconstruction Accuracy}. To address \textbf{Q1} and \textbf{Q3}, we conducted a comprehensive comparison of world state reconstruction metrics across different frameworks. We evaluated reconstruction accuracy using 1,000 episodes, with all policies trained under comparable reward structures and hyperparameters over 50,000 episodes. The reconstruction error is calculated as the average of the MSE of all metrics. As demonstrated in Fig.~\ref{fig:radar_plot}, the WMR framework consistently outperforms all baseline methods across all metrics. Notably, when compared to the variant without gradient cutoff, we observed a significant performance gap due to interference from competing gradients, which caused goal dispersion. The introduction of gradient cutoff improved reconstruction accuracy by approximately 40\%. Without Gradient Cutoff, the WMR fails to learn basic walking behaviors such as foot lifting and exhibits immediate instability when subjected to external perturbations or low-friction conditions, leading to premature episode termination. This behavioral failure partially explains the relatively smaller degradation observed in the accuracy metrics for contact mask and base velocity estimation. It is particularly noteworthy that the reconstruction of privileged information is achieved without direct supervision, relying solely on the estimator's ability to infer these states from historical observations. The WMR framework's superior performance in reconstructing these privileged states, despite this inherent challenge, provides compelling evidence of its architectural advantages over existing approaches.

\noindent\textbf{Comparisons on Episode Metrics}. To address \textbf{Q2} and \textbf{Q3}, we compared the robot's episode metrics across different frameworks. As shown in Table~\ref{tab:baselines}, the WMR framework outperforms all baselines in every metric. Specifically, compared to the method without reconstruction (\textbf{PPO}) and the method without gradient cutoff (\textbf{WMR without Gradient Cutoff, Denoising World Model Learning}), our approach with reconstruction excels in all episode metrics. Compared to the randomly sampled baseline (\textbf{WMR with Random Command Sample}), it has a slight increase in both mean velocity error and angular velocity error, demonstrating the advantage of training with motion capture data. These results validate the effectiveness of our framework.

\begin{figure}[t]
    \centering
    \includegraphics[width=0.489\textwidth]{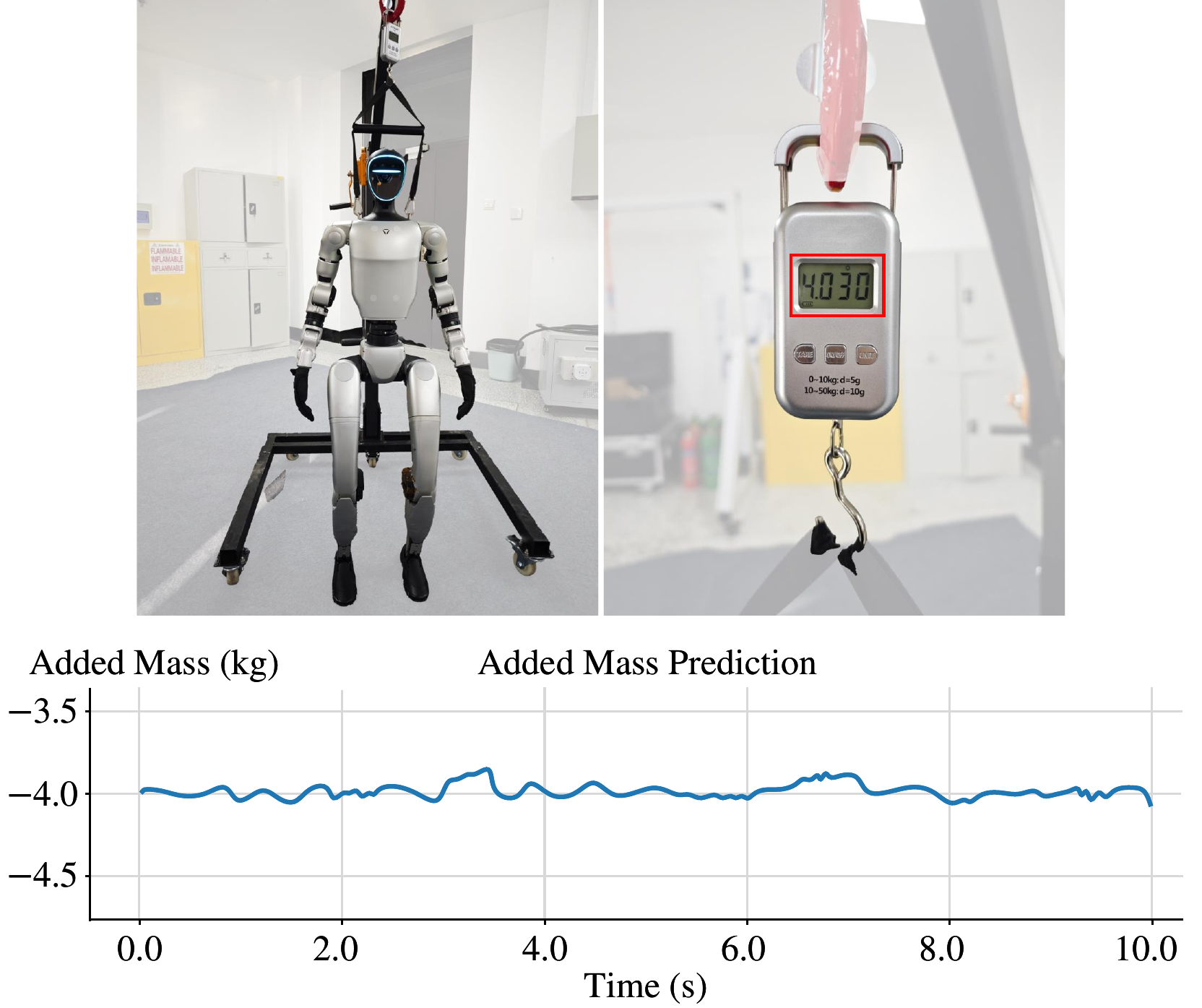}
    \caption{We evaluate the accuracy of payload estimation in the real-world. The actual unloaded weight is measured by a digital scale.}
    \label{fig:est_mass}
\end{figure}

\subsection{Ablation on Policy Training}
To respond to \textbf{Q4}, an ablation study was carried out on the WMR framework. This study focused on evaluating design decisions of the architectures. As shown in Table~\ref{tab:Ablation}, the LSTM, GRU, MLP, and Transformer architectures were compared for the estimator encoder, policy network, and value network. The LSTM and GRU architectures utilize hidden states to maintain memory over time, whereas the MLP and Transformer architectures rely on a fixed context window of the last 25 steps of observations. While the Transformer achieved the lowest $E_{\text{vel}}$, outperforming the LSTM-based WMR in velocity tracking, the LSTM architecture demonstrated superior performance in $E_{\text{ang}}$, $E_{\text{recon}}$, $M_{\text{terrain}}$, and $M_{\text{reward}}$. This discrepancy may be due to the limited of training episodes or the insufficient context window length, which might prevent the Transformer from fully leveraging its sequence modeling capabilities. The GRU demonstrated a slightly inferior performance compared to the LSTM, while the MLP performed the least satisfactorily. This is likely due to its limited ability to handle time series data.

\subsection{Sim-to-Real Evaluation}
To address the \textbf{Q1} and \textbf{Q5}, we evaluate the WMR framework on the sim and real-world terrain. The detailed experimental setup can be found in the Appendix~\hyperref[subsec:append_exp]{A}.

\begin{figure}[t]
    \centering
    \includegraphics[width=0.489\textwidth]{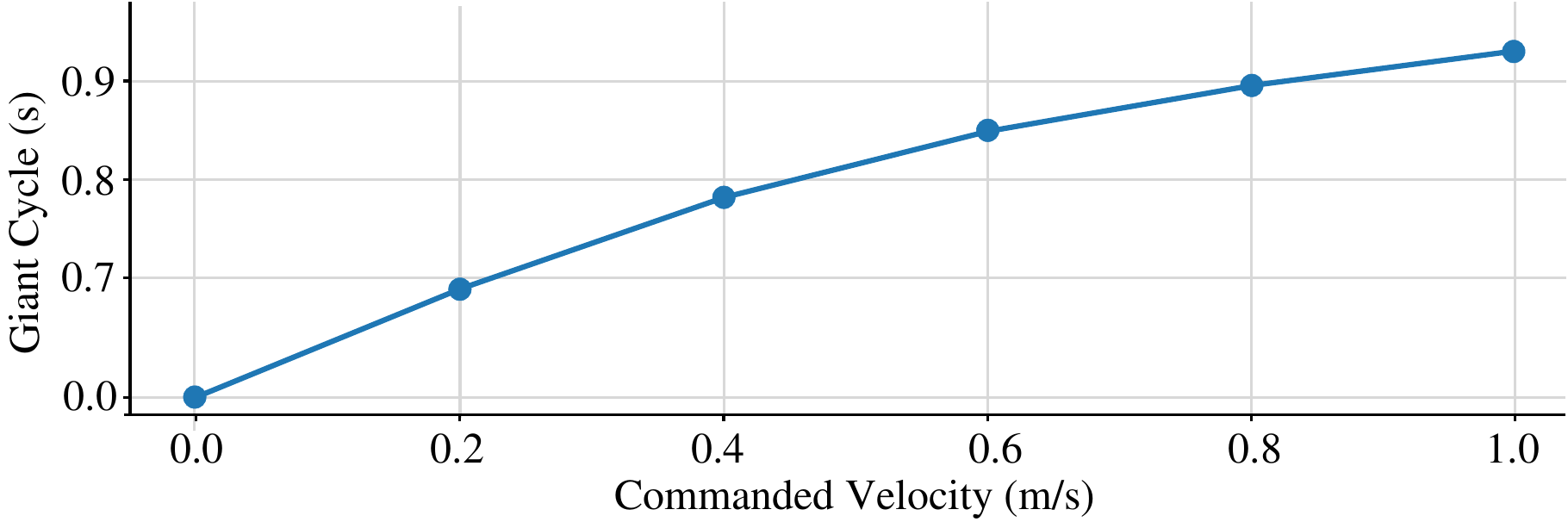}
    \caption{We compared the robot's walking cycles under different commands and found that the robot can adapt its step frequency according to the command.}
    \label{fig:gaint}
\end{figure}

\subsubsection{Real-World Deployment}
We present the results of deploying our controller in an outdoor environment. We conducted approximately 20 hours of testing over one week, encompassing various terrains and operational conditions. During this testing phase, the robot operated flawlessly without any failures, despite the absence of a gantry support system. The robot successfully navigated a maximum single distance of approximately 3.2 kilometers to reach the designated destination.

As shown in Figure~\ref{fig:head}, the robot exhibited robust mobility across terrains featuring low friction surfaces, deformable ground, and omnidirectional inclines. Throughout the deployment, human operators issued velocity commands to assess traversable areas, and the robot accurately followed these commands. Our findings indicate that the robot's capabilities closely align with human assessments of traversability, enabling the controller to reliably navigate all tested terrains without any physical assistance.

\begin{figure}[t]
    \centering
    \includegraphics[width=0.489\textwidth]{figures/walk.pdf}\\[6pt]
    \includegraphics[width=0.489\textwidth]{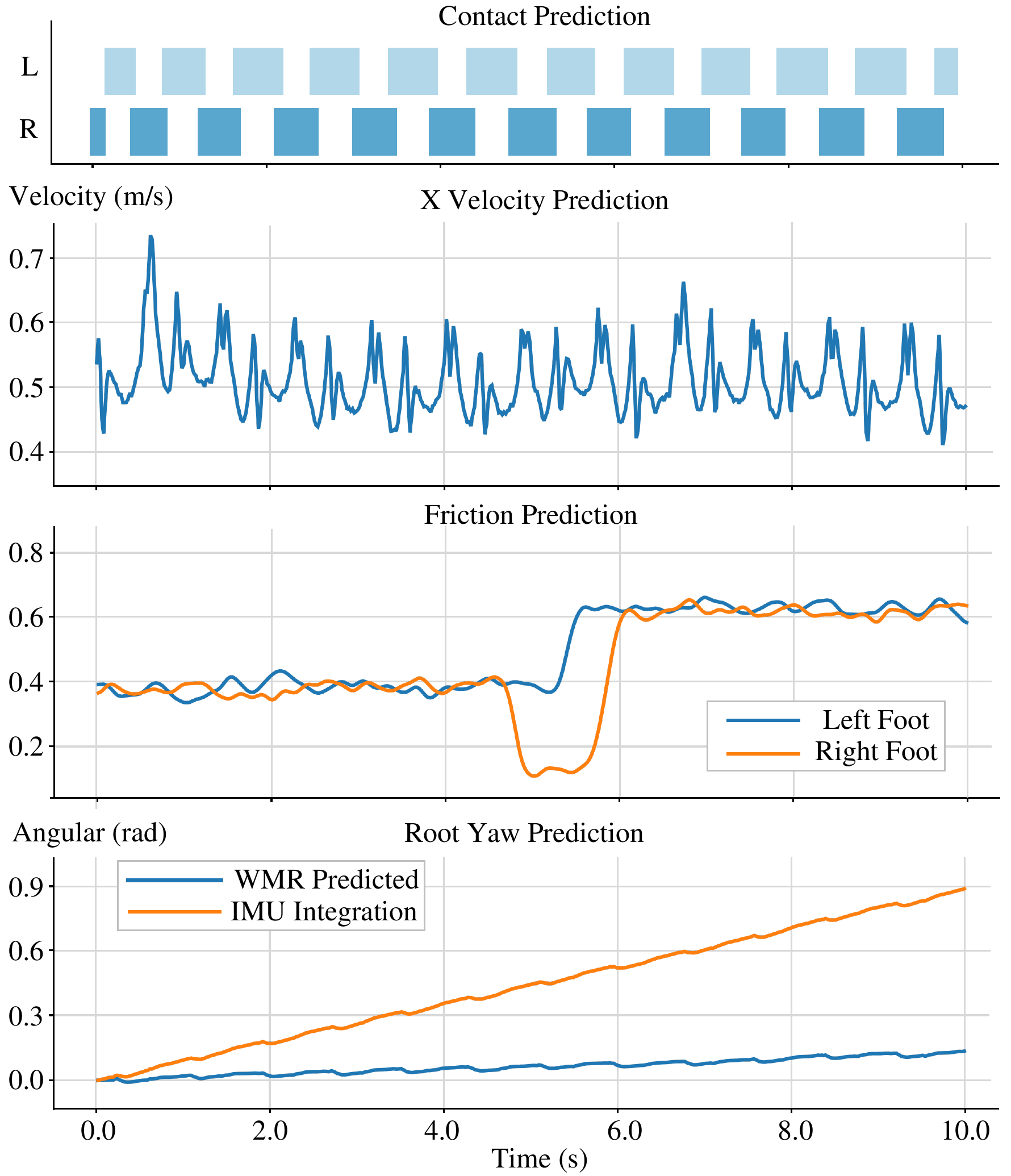}
    \caption{The robot walks steadily on surfaces of different materials. The robot is commanded to walk at a speed of 0.5 m/s in the $\mathit{x}$ direction.}
    \label{fig:walk}
\end{figure}

\subsubsection{Reconstruction Evaluation}

We conduct an evaluation of the WMR framework within the stair descent scenario. The goal is to determine its ability to accurately reconstruct in an environment that was not encountered during the training process.
The robot was instructed to proceed forward at a velocity of 0.3 m/s. Its path involved traversing flat ground, descending stairs, and then traversing flat ground again. This trajectory subjected the robot to a significant variation in the z-axis speed.
As depicted in Fig.~\ref{fig:est_sim}, \textbf{GT} represents the ground truth data, and \textbf{P} represents the WMR framework prediction. The WMR framework demonstrates the capacity to precisely reconstruct the world state. This includes elements such as the contact mask and linear velocity.

\subsubsection{Payload Estimation}
We evaluate the accuracy of added mass estimation in real-world, as shown in Fig.~\ref{fig:est_mass}. The robot was commanded to stand still in 10 seconds and given loads ranging from -10 kg to 10 kg. The actual loaded weight is determined by balancing the pre-calibrated loads on the front and back of the robot, while the actual unloaded weight is measured by an electronic scale. The average reconstruction accuracy of the 20 experiments is about 0.86.

\subsubsection{Gait Patterns Study}

One notable aspect of the reward function in the WMR framework is that it is deliberately designed without motion references. This design choice allows the robot to adjust its step frequency based on the magnitude of the commanded speed. As shown in Fig.~\ref{fig:gaint}, when the command speed is zero, the robot maintains a stable stance and stand still. When non zero speed instructions are given, it consistently tracks the speed. Additionally, as the command speed increases, the stepping cycle of both feet gradually rises, similar to the natural human walking patterns.

\subsubsection{Real-World Reconstruction}
We captured a 10-second segment during the deployment process and analyzed the reconstruction results, as shown in Fig.~\ref{fig:walk}. The robot was commanded to walk in the x direction at a speed of 0.5 m/s, involving three types of ground with different friction, namely, a brick ground with medium friction, a marble surface with very low friction, and an asphalt road with high friction. We observed that in the friction estimation, the WMR framework accurately estimates the friction trends of the three types of ground. At about 4.6 seconds, the right foot contacts the marble road, and the WMR responds quickly and estimates the decrease in ground friction, while the left foot is not affected. The estimated speed is very close to the command speed. Additionally, the estimated yaw angle is derived from the original sensor data and the integral of the angular velocity after noise reduction by the WMR framework. The WMR framework can effectively suppress the accumulated data drift caused by IMU noise.

\section{Discussions} 
\label{sec:conclusion}

In this work, we introduced the World Model Reconstruction (WMR) framework, a novel approach that enables robots to reconstruct world states using only their onboard sensors. A key innovation of our framework is the gradient cutoff mechanism, which significantly improves the accuracy of state estimation and the capability of the locomotion policy. By incorporating motion capture dataset, we substantially improved the robot's command tracking capabilities. Our extensive evaluations demonstrate that the WMR framework consistently outperforms existing baseline methods in both reconstruction accuracy and episode metrics, even effectively generalizing to previously unseen environments. The framework exhibited robust performance across challenging real-world conditions, including slopes, low-friction surfaces, deformable terrain, and rough ground, culminating in a successful 3.2km traversal without human intervention. Looking ahead, we plan to integrate environmental terrain perception into the WMR framework to tackle even more challenging terrains.

\section{Limitations}
One limitation of this study is that the robot cannot sense the shape of the terrain, which restricts the robot's mobility, especially on terrains that are challenging to navigate without height information, such as high platforms or discrete jumping obstacles. A potential solution to this limitation is to incorporate the height map as an input and utilize the WMR framework for denoising, thereby enhancing the robot's ability to traverse complex and demanding terrains.

Another limitation of this framework is that the input of the motion policy all comes from the reconstruction information. This may increase the risk of estimation errors causing system crash. Although we did not encounter the situation during the 20h deployment, it is a potential hidden danger, especially in situations the policy has never seen before.
\clearpage



\bibliographystyle{plainnat}
\bibliography{references}

\clearpage

\section*{Appendix}

\subsection{Experiment Setup}
\label{subsec:append_exp}
We conducted deployment experiments using the G1 humanoid robot. In the deployment, the model was executed in a Just-In-Time (JIT) mode with the C++ implementation of Onnx Runtime on the robot's onboard CPU, achieving an inference time of approximately 5 ms per inference. The model operated at an inference frequency of 50 Hz and communicated with the robot using the Data Distribution Service (DDS) at a frequency of 500 Hz. In simulation, we use IsaacLab~\cite{mittal2023orbit} for robot training and evaluation, with a policy inference frequency of 50 Hz.

\subsection{Domain Randomization}
\label{subsec:append_dr}
We use domain randomization to simulate the sensor noise in the real-world, the estimator reconstructs the world state from noisy observations. The randomization parameters are shown in Table~\ref{tab:domain_random}.

\begin{table}[h]
    \centering
    \small
    \begin{tabular}{l c c c}
        \toprule
        \textbf{Parameter} & \textbf{Unit} & \textbf{Range} & \textbf{Operator} \\
        \midrule
        Angular Velocity & rad/s & [-0.2, 0.2] & scaling \\
        Projected Gravity & - & [-0.1, 0.1] & scaling \\
        Joint Position & rad & [-0.1, 0.1] & scaling \\
        Joint Velocity & rad/s & [-1.5, 1.5] & scaling \\
        Friction Coefficient & - & [0.2, 1.5] & - \\
        Payload & kg & [-5.0, 5.0] & additive \\
        Gravity & m/s$^2$ & [-0.1, 0.1] & additive \\
        Joint Damping & - & [0.8, 1.2] & scaling \\
        Joint Stiffness & - & [0.8, 1.2] & scaling \\
        Motor Offset & rad & [-0.1, 0.1] & additive \\
        \bottomrule
    \end{tabular}
    \caption{Domain randomization parameters. Additive randomization adds a random value within a specified range to the parameter, while scaling randomization adjusts the parameter by a random multiplication factor within the range.}
    \label{tab:domain_random}
\end{table}

\subsection{Reward Function}
\label{subsec:append_reward}
Our reward function is a sum of the following terms:

\begin{itemize}

\item \textbf{Tracking Linear Velocity Reward} ($r_{\text{vel}}$): This term encourages the robot to track the commanded linear velocity in the $\mathit{xy}$-plane.
\begin{equation*}
    r_{vel} := \exp(-\|v_{xy} - v_{xy}^*\|_2^2/\sigma_{vel}),
\end{equation*}
where $v_{xy}$ and $v_{xy}^*$ represent the actual and commanded linear velocities, respectively. $\sigma_{vel}$ is set to 0.25.

\item \textbf{Tracking Angular Velocity Reward} ($r_{\text{ang}}$): This term encourages the robot to track the commanded angular velocity.
\begin{equation*}
    r_{ang} := \exp(-\|\omega - \omega^*\|_2^2/\sigma_{ang}),
\end{equation*}
where $\omega$ and $\omega^*$ represent the actual and commanded angular velocities, respectively. $\sigma_{ang}$ is set to 0.25.
        
\item \textbf{Termination Reward}: This term penalizes episode termination.
\begin{equation*}
    r_{ter} := -200.0 * \mathbb{I}_{\text{terminated}}
\end{equation*}
where $\mathbb{I}_{\text{terminated}}$ is 1 if the episode terminates, otherwise 0.

\item \textbf{Z-axis Linear Velocity Reward}: This term penalizes the robot for moving along the z-axis.
\begin{equation*}
    r_{z} := -1.0 * (v_z)^2
\end{equation*}
where $z$ is the z-axis linear velocity.

\item \textbf{Energy Reward}: This term penalizes output torques to reduce energy consumption.
\begin{equation*}
    r_{e} := -0.001 * \sum_i |\tau_i \cdot \dot{q}_i|
\end{equation*}
where $\tau$ represents the joint torques and $\dot{q}$ represents the joint velocities.

\item \textbf{Base Angular Velocity Reward}: This term penalizes excessive angular velocity in the $\mathit{xy}$-plane to maintain stability.
\begin{equation*}
    r_{ang} := -0.05 * \|\omega_{xy}\|_2^2
\end{equation*}
where $\omega_{xy}$ represents the base angular velocity in the $\mathit{xy}$-plane.

\item \textbf{Joint Acceleration Reward}: This term penalizes excessive joint accelerations to promote smooth motions.
\begin{equation*}
    r_{ja} := -2.5* 10^{-7} * \|\ddot{q}\|_2^2
\end{equation*}
where $\ddot{q}$ represents the joint accelerations of the configured joints.

\item \textbf{Action Rate Reward}: This term penalizes rapid changes in actions to encourage smooth control.
\begin{equation*}
    r_{ar} := -0.01 * \|a_t - a_{t-1}\|_2^2
\end{equation*}
where $a_t$ represents the current action and $a_{t-1}$ represents the previous action.

\item \textbf{Base Orientation Reward}: This term penalizes non-flat base orientation to maintain an upright posture.
\begin{equation*}
    r_{ori} := -2.0 * \|g_{xy}\|_2^2
\end{equation*}
where $g_{xy}$ represents the $\mathit{xy}$-components of the projected gravity vector in the base frame. A perfectly upright orientation would have zero $\mathit{xy}$-components.

\item \textbf{Joint Position Limit Reward}: This term penalizes joint positions that exceed their soft limits.
\begin{equation*}
    r_{\text{jpl}} := -2.0 \cdot \sum_i \bigl[ \max(q_i{-}q_{i,\max},0) + \max(q_{i,\min}{-}q_i,0) \bigr]
\end{equation*}
where $q_i$ represents the position of joint $i$, and $q_{i,\min}$ and $q_{i,\max}$ are the lower and upper soft limits for that joint, respectively.

\item \textbf{Joint Deviation Reward}: This term penalizes joint positions that deviate from their default positions.
\begin{equation*}
\begin{aligned}
    r_{jd} := &-0.05 * \sum_i |q_i - q_{i,default}| \\
              &-0.1 * \sum_j |q_j - q_{j,default}| \\
              &-0.2 * \sum_k |q_k - q_{k,default}|
\end{aligned}
\end{equation*}
where $q$ represents the position of a joint, and $q_{default}$ is the default position for that joint. In our study, $i$ represents the hip pitch joint, knee pitch joint, and ankle pitch joint; $j$ represents the hip yaw joint and hip roll joint; and $k$ represents all other joints.

\item \textbf{Feet Air Time Reward}: This term rewards appropriate stepping behavior for bipedal locomotion.
\begin{equation*}
    r_{fat} := \begin{cases}
        0.2 * \min(t, \theta) & \text{if } \sum_{i} c_i = 1 \text{ and } \|v\| > 0.1 \\
        0 & \text{otherwise}
    \end{cases}
\end{equation*}
where $t$ is either the contact time or air time for each foot (whichever is active), $c_i$ is the contact state of foot $i$, $\theta$ is set to 0.4 $\mathrm{s}$, and $v$ is the commanded velocity.

\item \textbf{Feet Force Reward}: This term encourages maintaining appropriate ground reaction forces.
\begin{equation*}
    r_{ff} := \begin{cases}
        5 * 10^{-3} * \min(f_z - f_{th}, f_{max}) & \text{if } f_z > f_{th} \\
        0 & \text{otherwise}
    \end{cases}
\end{equation*}
where $f_z$ is the vertical ground reaction force, $f_{th}=500N$ is the threshold force, and $f_{max}=400N$ is the maximum reward.

\item \textbf{Feet Stumble Reward}: This term penalizes lateral forces that indicate stumbling.
\begin{equation*}
    r_{fs} := -2.0 * \mathbb{I}(\|f_{xy}\|_2 > f_z)
\end{equation*}
where $f_{xy}$ represents the horizontal ground reaction forces and $f_z$ represents the vertical ground reaction force.

\item \textbf{Feet Sliding Reward}: This term penalizes feet sliding during ground contact.
\begin{equation*}
    r_{sl} := -0.25 * \sum_i \|v_{i,xy}\|_2 \cdot \mathbb{I}(f_i > 1)
\end{equation*}
where $v_{i,xy}$ is the horizontal velocity of foot $i$, and $\mathbb{I}(f_i > 1)$ indicates if the foot is in contact with the ground.

\item \textbf{Flying State Reward}: This term penalizes the robot when it is airborne.
\begin{equation*}
    r_{fly} := -\mathbb{I}(\sum_i t_{c,i} < \epsilon)
\end{equation*}
where $t_{c,i}$ is the contact time of foot $i$ and $\epsilon = 0.001s$.

\item \textbf{Undesired Contacts Reward}: This term penalizes undesired contacts with the environment.
\begin{equation*}
    r_{uc} := -\sum_{i \in \mathcal{C}} \mathbb{I}\left( \| \mathbf{F}_i \|_2 > 1.0 \right)
\end{equation*}
where $\mathcal{C}$ represents the set of contact points excluding those involving the ankle. The indicator function $\mathbb{I}$ takes the value 1 when the contact force $\| \mathbf{F}_i \|_2$ exceeds the threshold of 1.0 $\mathrm{N}$, and 0 otherwise.

\end{itemize}

\subsection{WMR Hyperparameters}
We illustrate the hyperparameters of WMR in Table.~\ref{tab:hyperparameters}.

\begin{table}[h]
    \centering
    \small

    \begin{tabular}{l r}
        \toprule
        \textbf{Parameter} & \textbf{Value} \\
        \midrule
        Number of Environments & 4096 \\
        Training Iteration & 100000 \\
        Environment Steps & 24 \\
        Number of Training Epochs & 5 \\
        Mini Batch Size & 24576 \\
        Max Clip Value Loss & 0.2 \\
        Discount Factor & 0.99 \\
        GAE discount factor & 0.95 \\
        Entropy Regularization Coefficient & 0.01 \\
        Learning rate & 2.5e-5 \\
        $\lambda_{\mathrm{cont}}$ & 1.0 \\
        $\lambda_{\mathrm{dis}}$ & 0.3 \\
        $\lambda_{\mathrm{reg}}$ & 0.005 \\
        $\lambda_v$ & 1.0 \\
        $\lambda_\pi$ & 1.0 \\
        \bottomrule
    \end{tabular}
    \caption{Hyperparameters of WMR.}
    \label{tab:hyperparameters}
\end{table}

\subsection{Architecture Details}

We illustrate the network architecture of WMR in Table.~\ref{tab:network_arch}.

\begin{table}[h]
    \centering
    \small
    \begin{tabular}{|l|c|}
        \toprule
        Component & Configuration \\
        \midrule
        \multicolumn{2}{|c|}{History Encoder} \\
        \midrule
        RNN Memory (0) & LSTM(96 → 256) \\
        \midrule
        \multicolumn{2}{|c|}{Continuous Decoder} \\
        \midrule
        Con. Decoder (0) & ELU(alpha=1.0) \\        
        Con. Decoder (1) & Linear(256 → 256) \\
        Con. Decoder (2) & ELU(alpha=1.0) \\
        Con. Decoder (3) & Linear(256 → 190) \\
        \midrule
        \multicolumn{2}{|c|}{Discrete Decoder} \\
        \midrule
        Dis. Decoder (0) & ELU(alpha=1.0) \\
        Dis. Decoder (1) & Linear(256 → 256) \\
        Dis. Decoder (2) & ELU(alpha=1.0) \\
        Dis. Decoder (3) & Linear(256 → 2) \\
        Dis. Decoder (4) & Sigmoid() \\
        \midrule
        \multicolumn{2}{|c|}{Locomotion Policy} \\
        \midrule
        RNN Memory (0) & LSTM(192 → 256) \\
        Policy Net (1) & ELU(alpha=1.0) \\ 
        Policy Net (2) & Linear(256 → 256) \\
        Policy Net (3) & ELU(alpha=1.0) \\
        Policy Net (4) & Linear(256 → 128) \\
        Policy Net (5) & ELU(alpha=1.0) \\
        Policy Net (6) & Linear(128 → 29) \\
        \midrule
        \multicolumn{2}{|c|}{Value Network} \\
        \midrule
        RNN Memory (0) & LSTM(192 → 256) \\
        Value Net (1) & ELU(alpha=1.0) \\ 
        Value Net (2) & Linear(256 → 256) \\
        Value Net (3) & ELU(alpha=1.0) \\
        Value Net (4) & Linear(256 → 128) \\
        Value Net (5) & ELU(alpha=1.0) \\
        Value Net (6) & Linear(128 → 1) \\
        \bottomrule
    \end{tabular}
    \caption{World Model Reconstruction Network architecture details. The table shows the configuration of each component including encoder, decoder, locomotion policy and value network.}
    \label{tab:network_arch}
\end{table}

\end{document}